\pdfoutput=1
\documentclass[11pt]{article}

\usepackage[final]{acl}

\usepackage{times}
\usepackage{latexsym}

\usepackage[T1]{fontenc}
\usepackage[utf8]{inputenc}

\usepackage{microtype}

\usepackage{inconsolata}

\usepackage{amsmath}
\usepackage{multirow}
\usepackage{listings}
\usepackage{color}
\usepackage{graphicx}
\usepackage{subcaption}
\usepackage{makecell}
\usepackage{booktabs}
\usepackage{cleveref}
\usepackage{colortbl}
\usepackage{soul}
\usepackage{xcolor}
\usepackage{empheq}
\usepackage[many]{tcolorbox}
\usepackage{stfloats}
\usepackage{bbding}
\usepackage{tabularx}

\definecolor{dkgreen}{rgb}{0,0.6,0}
\definecolor{gray}{rgb}{0.5,0.5,0.5}
\definecolor{mauve}{rgb}{0.58,0,0.82}

\definecolor{dgreen}{rgb}{0.412,0.741,0.271}
\definecolor{dblue}{rgb}{0.220,0.325,0.639}
\definecolor{dred}{rgb}{0.933,0.122,0.137}

\definecolor{g1}{HTML}{b3e2cd}
\definecolor{r1}{HTML}{fdcdac}
\definecolor{w1}{HTML}{cbd5e8}
\definecolor{b1}{HTML}{fff7bc}

\definecolor{lr}{HTML}{bebada}
\definecolor{fr}{HTML}{fccde5}

\definecolor{Lavender}{HTML}{BF94E4}

\newcommand{\ie}{\textit{i}.\textit{e}.\,}
\newcommand{\eg}{\textit{e}.\textit{g}.\,}

\newcommand{\centerone}[2]{\multicolumn{1}{>{\columncolor{#1}}c|}{#2}}

\newcommand{\centeronerdeep}[1]{\centerone{r1!150}{#1}}
\newcommand{\centeronerlight}[1]{\centerone{r1!50}{#1}}

\newcommand{\centeronewdeep}[1]{\centerone{w1!170}{#1}}
\newcommand{\centeronewlight}[1]{\centerone{w1!70}{#1}}

\newcommand{\centeronenobar}[2]{\multicolumn{1}{>{\columncolor{#1}}c}{#2}}

\newcommand{\centeronerdeepnobar}[1]{\centeronenobar{r1!150}{#1}}
\newcommand{\centeronerlightnobar}[1]{\centeronenobar{r1!50}{#1}}
\newcommand{\centeronewdeepnobar}[1]{\centeronenobar{w1!170}{#1}}
\newcommand{\centeronewlightnobar}[1]{\centeronenobar{w1!70}{#1}}

\definecolor{l1}{RGB}{189,215,238}
\definecolor{l2}{RGB}{222,235,247}
\definecolor{l3}{RGB}{255,230,153}
\definecolor{l4}{RGB}{248,203,173}
\definecolor{l5}{RGB}{244,177,131}

\newtcbox{\columntcbox}{highlight math style={
        colback=gray!30,
        arc=2pt,
        outer arc=2pt,
        boxrule=0pt,
        top=2pt,
        bottom=2pt,
        left=2pt,
        right=2pt,
    }
}

\colorlet{LightLavender}{Lavender!35!}
\newtcbox{\inlinetcbox}[1][]{on line, 
        boxsep=2pt, left=0pt,right=0pt,top=2pt,bottom=-1pt,
        colframe=white,colback=LightLavender,  
        highlight math style={enhanced}, #1
}

\newcommand\blfootnote[1]{%
  \begingroup
  \renewcommand\thefootnote{}\footnote{#1}%
  \addtocounter{footnote}{-1}%
  \endgroup
}

\newcolumntype{C}[1]{>{\centering}m{#1}}

\title{\texttt{FAC$^2$E}: Better Understanding Large Language Model Capabilities by Dissociating Language and Cognition}

\author{
    Xiaoqiang Wang\textsuperscript{\rm 1,2},
    Lingfei Wu\textsuperscript{\rm 3},
    Tengfei Ma\textsuperscript{\rm 4}
    \and Bang Liu\textsuperscript{\rm 1,2$\dagger$} \\
    \textsuperscript{\rm 1}DIRO \& Institut Courtois, Universit{\'e} de Montr{\'e}al\\
    \textsuperscript{\rm 2}Mila - Quebec AI Institute;
    \textsuperscript{\rm 3}Anytime.AI;
    \textsuperscript{\rm 4}Stony Brook University \\
    \{\texttt{xiaoqiang.wang, bang.liu}\}\texttt{@umontreal.ca} \\
    \texttt{lwu@anytime-ai.com, tengfei.ma@stonybrook.edu}
}
\begin{document}
\maketitle
\begin{abstract}
Large language models (LLMs) are primarily evaluated by overall performance on various text understanding and generation \textit{tasks}.
However, such a paradigm fails to comprehensively differentiate the fine-grained language and cognitive \textit{skills}, rendering the lack of sufficient interpretation to LLMs' \textit{capabilities}. 
In this paper, we present \textbf{\texttt{FAC$^2$E}}, a framework for \underline{\textbf{F}}ine-gr\underline{\textbf{A}}ined and \underline{\textbf{C}}ognition-grounded LLMs' \underline{\textbf{C}}apability \underline{\textbf{E}}valuation.
Specifically, we formulate LLMs' evaluation in a multi-dimensional and explainable manner by dissociating the language-related capabilities and the cognition-related ones.
Besides, through extracting the intermediate reasoning from LLMs, we further break down the process of applying a specific capability into three sub-steps: recalling relevant knowledge, utilizing knowledge, and solving problems.
Finally, \texttt{FAC$^2$E} evaluates each sub-step of each fine-grained capability, providing a \underline{\textbf{two}}-faceted diagnosis for LLMs.
Utilizing \texttt{FAC$^2$E}, we identify a common shortfall in knowledge utilization among models and propose a straightforward, knowledge-enhanced method to mitigate this issue. Our results not only showcase promising performance enhancements but also highlight a direction for future LLM advancements.
\end{abstract}

\blfootnote{$^\dagger$Corresponding author. Canada CIFAR AI Chair.}
%\blfootnote{$^\ddagger$Corresponding authors.}

\section{Introduction}
\label{sec:introduction}

Large language models (LLMs)~\cite{brown2020language}, especially instruction-tuned LLMs~\cite{ouyang2022training, bai2022training, touvron2023llama2, chiang2023vicuna} revolutionized natural language processing and have 
surpassed human performance on tasks that require nontrivial reasoning~\cite{guo2023close, malinka2023educational}, while showing great potential in applications from conversational assistants~\cite{chatgpt, achiam2023gpt} to expertise problem-solving~\cite{nori2023capabilities, zhou2023solving, suzgun2024meta}.
%%
% However, despite the impressive performance, LLMs also show some notorious biases and hallucinations, \ie favoring certain positions over others~\cite{zheng2023judging} and generating content that is inconsistent with real-world facts or user inputs~\cite{rawte2023survey}.
However, despite the impressive performance, LLMs also show poor robustness on complex tasks~\cite{ullman2023large} and significantly inconsistent evaluation results under different settings, such as binary preference~\cite{xu2023wizardlm} and automatic metrics~\cite{gudibande2023false}.
Therefore, it is crucial to attain an overarching understanding of the capabilities and limitations of LLMs.

%{\color{dred}
To address this challenge, some studies have assessed the performance of LLMs on different tasks based on independent benchmarks from various dimensions~\cite{liang2022holistic, srivastava2023beyond,eval-harness}, such as commensense~\cite{zellers-etal-2019-hellaswag}, knowledge~\cite{hendrycks2020measuring,yu2023kola}, instruction-following~\cite{gu2024dingo}, and trustworthy~\cite{sun2024trustllm}.
%, and hallucination~\cite{li-etal-2023-halueval}.
% and alignments to human values~\cite{chia2023instructeval}.
% based on automatic metrics~\cite{longpre2023flan}, teacher models~\cite{wang2023far, zheng2023judging}, or human preferences~\cite{xu2023wizardlm}.
%%
Besides, motivated by building LLM-based AI assistants, other studies propose highly curated benchmarks with instance-level fine-grained annotations, such as difficulty and reasoning skills~\cite{mialon2023gaia, ye2024flask}, for holistic evaluation of LLMs.
% The others devise a suite of dedicated tasks to evaluate the LLMs' flexibility to adapt learned capability to novel task variants~\cite{wu2023reasoning}.

% However, the existing studies overlook a crucial distinction between 
% \textit{language} and \textit{cognition}~\cite{monti2012thought, blank2014functional}, rendering the insufficient understanding of a model's true capabilities and effectiveness in tasks.
%%

However, the existing studies, assessing effectiveness across various tasks, provided limited insight into the models' true capabilities, as it only indicates their overall performance on specific datasets, without revealing the fine-grained capabilities acquired or their proficiency levels within the multiple capabilities involved.
For instance, in the context of generative question answering, a model adept at extracting information but struggling to form a coherent understanding may exhibit similar overall performance to another model with profound understanding insights but difficulties in articulating accurate responses.

We argue that a fine-grained understanding of LLMs' capabilities can not only accurately unveil their inherent limitations, but also help us to better identify why one model outperforms the other and how the different capabilities correlate.
% with each other.
%%
Additionally, such insights allow us to provide tailored guidance to improve training efficiency or facilitate more advanced model development.
%}

% the possible reasons behind that
% akin to a reader reciting without true comprehension. Conversely, imagine a model with deep cognitive insights but challenges in articulating accurate responses, resembling a thinker struggling to convey insights coherently
% %%
% What is worse, they quantify each capability only using the performance metric on test tasks, hence conveying less about why one model outperforms the other and how the different capabilities correlate with each other.
%% rendering the insufficiency in (i) comprehensiveness of capabilities coverage and (ii) interpretability of evaluation.

% We argue that a multi-dimensional and interpretable understanding of LLMs' capabilities can not only accurately unveil their inherent limitations, but also help us to identify the possible reasons behind that, 
% % such as the quality of fine-tuning datasets and the superiority of certain backbone models, 
% such as the quality of \red{knowledge store} and efficacy of knowledge utilization, 
% further allow us to provide tailored guidance to improve training efficiency or facilitate more advanced model development.

In this paper, we propose \texttt{FAC$^2$E}, a fine-grained capability evaluation framework for LLMs.
Specifically, \texttt{FAC$^2$E} dissociates the language-related and cognition-related capabilities of LLMs and organizing them into four distinct axes: \textsc{Linguistic Knowledge}, \textsc{Formal Knowledge}, \textsc{World Modeling}, and \textsc{Social Modeling}.
This categorization is grounded in neuroscience evidence manifesting that language processing and cognitive processes, like memory and reasoning, operate differently in the brain.
% which are inspired by a wealth of empirical evidence from neuroscience.
% %%
% Brain imaging studies manifest that linguistic mechanisms are separated from cognitive functions, such as memory and reasoning.
%%
Drawing from this insight, we adapt a range of existing benchmarks into a unified question-answering format.
We then develop specific instructions for each capability, allowing \texttt{FAC$^2$E} to evaluate LLMs through a method known as few-shot instruction-following.
% %%
% Based on this, we leverage various existing benchmarks and unify them into a question-answering format.
% %%
% Then, we design capability-specific instructions and formulate \texttt{FAC$^2$E} via few-shot instruction-following.

Furthermore, we break down the application of a specific capability into three sub-steps: knowledge recall, knowledge utilization, and problem-solving, by iteratively drawing out the model's intermediate reasoning.
After evaluating each sub-step, \texttt{FAC$^2$E} can reveal the quality of knowledge encoded in the model, and effectiveness in applying relevant knowledge to solve practical problems, offering a more comprehensive evaluation than a single performance metric could.

Our findings reveal a notable gap in capabilities between open-source and proprietary models, especially for cognition-related capabilities.
Additionally, we found that many models have difficulties in applying knowledge effectively.
To address this, we suggest a knowledge-enhance remedy by incorporating relevant knowledge text as additional input. 
Experimental results show that it can help the backbone model (\eg LLaMA 2) achieve approximately 90\% of the performance of its instruction-tuned counterpart (\eg LLaMA 2-Chat).

\begin{table*}[t!]
    \resizebox{1.0\textwidth}{!}{
        \begin{tabular}{clll}
            \toprule
            \multicolumn{1}{c}{\textbf{Capability}}& \multicolumn{1}{c}{\textbf{Description}}& \multicolumn{2}{c}{\textbf{Skill Example}} \\
            \hline
            \multirow{3}{2.5cm}{\centering \textbf{\textsc{Linguistic Knowledge}}}& \cellcolor{l2}& \cellcolor{l2}&  \cellcolor{l2} \\
            & \cellcolor{l2} & \multirow{-2}{*}{\cellcolor{l2}Grammaticality:} & \multirow{-2}{8cm}{\cellcolor{l2}agreements, licensing, long-distance dependencies, and garden-path effects.}\\
            & \multirow{-3}{7cm}{\cellcolor{l2}Encoding grammatical concepts support linguistic operations regarding word meanings and their combinatorial processing.}& \cellcolor{l2}Semantics:& \cellcolor{l2}synonymy, antonymy, and hypernymy. \\
            \hline
            \multirow{2}{2.5cm}{\centering \textbf{\textsc{Formal Knowledge}}}& \cellcolor{l1}& \cellcolor{l1}Mechanism:& \cellcolor{l1}deductive, inductive, and analogical. \\
            & \multirow{-2}{7cm}{\cellcolor{l1}Conducting word-based formal reasoning through understanding lexical semantics.}& \cellcolor{l1}Skill:& \cellcolor{l1}numeric, logic, and manipulation. \\
            \hline
            \multirow{2}{2.5cm}{\centering \textbf{\textsc{World Modeling}}}& \cellcolor{l4}& \cellcolor{l4}Remember:& \cellcolor{l4}factual knowledge, context, and commensense. \\
            & \multirow{-2}{7cm}{\cellcolor{l4}Understanding text based on given context and associating it with world knowledge.}& \cellcolor{l4}Understand:& \cellcolor{l4}narrative structure and discourse comprehension. \\
            \hline
            \multirow{3}{2.5cm}{\centering \textbf{\textsc{Social Modeling}}}& \cellcolor{l5} & \cellcolor{l5} & \cellcolor{l5} \\ 
            & \cellcolor{l5} & \multirow{-2}{*}{\cellcolor{l5}Pragmatics:} & \multirow{-2}{8cm}{\cellcolor{l5}polite deceits, irony, maxims of conversation, metaphor, indirect speech, and humor.}\\
            & \multirow{-3}{7cm}{\cellcolor{l5}Infering mental state behind text and intended meaning beyond literal content.}& \cellcolor{l5}Theory-of-mind& \cellcolor{l5}unexpected content and unexpected transfer tasks. \\
            \bottomrule
        \end{tabular}
        }
    \caption{Formulation of cognition-grounded LLMs' capabilities. See Section~\ref{subsec:formulation-of-capability} for details.}
    \label{tab:capability-schema}
    \vspace{-5mm}
\end{table*}

\section{Methodology}
\label{sec:methodology}

In this section, we introduce \texttt{FAC$^2$E} framework, designed for fine-grained and cognition-grounded LLMs' capability evaluation.
Specifically, we first define the taxonomy for LLMs' capabilities based on the distinction between language and cognition, which is drawn upon insights from neuroscience~\cite{fedorenko2016language, mahowald2023dissociating}.
Based on this, we transform a variety of existing benchmarks into the unified question-answering format, design capability-specific instruction, and frame \texttt{FAC$^2$E} via few-shot instruction-following.
Furthermore, we break down the evaluation process for each capability into a three-step reasoning approach.
This involves identifying the knowledge pertinent to the input, examining how the model applies this knowledge in practical contexts, and assessing the effectiveness of its problem-solving.
%%
% eliciting what knowledge the input is about, how the model adopts relevant knowledge into practical skills, and how good the problem-solving performance is.
%%
By evaluating each of these steps, \texttt{FAC$^2$E} provides a comprehensive overview of the model's performance, offering a more nuanced understanding of LLMs' intrinsic capabilities.
% Finally, through evaluating each sub-step, \texttt{FAC$^2$E} outputs an overall assessment for the tested model to better describe LLMs' inherent capabilities. 

\subsection{Formulation of LLMs' Capabilities}
\label{subsec:formulation-of-capability}
%{\color{dred}
Human language processing, long studied in cognitive science and neuroscience, robustly attributes language and cognition to different brain areas, namely ``language network'' and ``multi-demand network''~\cite{duncan2010multiple, scott2017new}. The former is sensitive to linguistic regularities and formal operations, with damage leading to linguistic deficits, while the latter responds actively to various cognitively demanding processes, such as reasoning and memory.
Similarly, prior analyses have identified core linguistic regions~\cite{zhang2024unveiling} and language-independent knowledge neurons~\cite{chen2023journey} in LLMs, represented by different parameter sets and subnetworks, each contributing distinctly to language and reasoning tasks.

Motivated by this separated relationship, we define the LLMs' capabilities as a 4-dimensional schema as shown in Table~\ref{tab:capability-schema}.
Compared to the fine-grained definitions of high-level knowledge~\cite{hendrycks2020measuring,yu2023kola} and reasoning skills~\cite{ye2024flask} in related works, this formulation aims to dissociate language-related and cognition-related capabilities to define a broader range of both high-level and more fundamental low-level functionality. Additionally, it seeks to minimize the coupling between these capabilities to facilitate nuanced analysis (Sec.~\ref{subsec:main-results}) and targeted improvement of the model (Sec.~\ref{subsec:booting-llms}).
%%
% We then proceed to explain each capability and underline the specific skills they encompass.
%}

\noindent
\textbf{\textsc{Linguistic Knowledge}.}\
To effectively generate language, LLMs first understand text at a basic linguistic level, including both grammar and semantics.
%%
% Specifically, on the one hand, grammaticality needs to be represented at different levels, varying from phonological, lexical, and word levels, to phrase and sentence levels.
% %%
% To embody the grammatical hierarchy as much as possible, based on the challenging tasks and inferior performance of conventional models in BLiMP benchmark~\cite{warstadt-etal-2020-blimp-benchmark}, we exemplify four skills, consisting of agreements (anaphor and subject-verb), licensing (negative polarity item and reflexive pronoun), long-distance dependencies (filler-gap and cleft), and garden-path effects.
%%
Grammaticality encompasses the rules that govern language structure, spanning from the sounds (phonological) and words (lexical) to the arrangement of words in phrases and sentences. To capture this grammatical structure as comprehensively as possible, especially given the challenges conventional models face in benchmarks like BLiMP~\cite{warstadt-etal-2020-blimp-benchmark}, we focus on four key skills: agreement (anaphora and subject-verb relationships), licensing (negative polarity items and reflexive pronouns), managing long-distance dependencies (filler-gap constructions and cleft sentences), and navigating garden-path sentences, which contain temporary ambiguities that must be resolved for correct understanding. For example, ``\emph{the horse raced past the barn fell,}” the initial interpretation is that the horse is racing, but upon reaching ``\emph{fell,}'' it becomes clear that the horse is being raced by another (unnamed) entity.

% On the other hand, although semantics is something closer to understanding, \ie high-level cognition, \textsc{Linguistic Knowledge} is associated with the understanding of word meanings, \ie lexical semantics, and irrelevant to general conceptual knowledge which can be categorized as other dimensions.
Semantics, on the other hand, while more closely related to high-level cognitive understanding, within the context of \textsc{Linguistic Knowledge}, pertains to the meanings of individual words or lexical semantics~\cite{geeraerts2009theories}. This aspect is distinct from conceptual knowledge, which falls under other dimensions of LLM capabilities, highlighting the meaning-related understanding, such as synonymy, antonymy, and hypernymy.

\noindent
\textbf{\textsc{Formal Knowledge}.}\
Beyond encoding linguistic structures and word meanings, an essential aspect of language capability involves understanding formal operations among words, or word-based reasoning. This means LLMs should be capable of recognizing relationships between words and deducing missing elements in a given pattern, such as completing analogies (\eg ``\textit{man:woman :: king:\_}'').
\texttt{FAC$^2$E} includes three types of reasoning mechanisms—deductive, inductive, and analogical reasoning—between words~\cite{bang-etal-2023-multitask}, and includes three symbol-based formal skills: numeric (dealing with numbers), logic (applying logical operations), and manipulation (altering the inputs in a rule-based manner)~\cite{wei2022chain}. 
An example task is concatenating the last letters of a word list (``\emph{think, machine, learning}'' $\rightarrow$ ``\emph{keg}'').

% For example, the last-letter-concatenation task~\cite{wei2022chain} where the input is a list of words, and the output is the concatenation of the last letters of the words in the list (``\emph{think, machine, learning}'' $\rightarrow$ ``\emph{keg}'').

\noindent
\textbf{\textsc{World Modeling}.}\
To step towards cognitive capabilities, well-grounded comprehension of factual and commonsense knowledge is required.
Precisely, we decompose this capability into two primary mechanisms: \textit{remember} and \textit{understand}, respectively modeling the retrieval-based and comprehension-based capability~\cite{sugawara2020assessing}.
Considering the versatility of knowledge sources, we instantiate the \textit{remember} sub-capability as recalling factual knowledge (open-ended facts), reading comprehension (facts in context), and applying commonsense reasoning.
Based on the multiple granularities of text comprehension and hierarchy of input text, we characterize the \textit{understand} sub-capability as two skills: understanding narrative or event structure (paragraph-level), and discourse comprehension (document-level).

\noindent
\textbf{\textsc{Social Modeling}.} \
The utility of human language lies in not only the understanding of the text itself but also the social context and mental states underlying communication~\cite{adolphs2009social}, \ie serving as a medium for information exchange between individuals.
%%
% As suggested by research of cognitive science~\cite{adolphs2009social}, the human brain has dedicated machinery, namely ``theory of mind network'', for processing social information and understanding somebody's mental state.
%%
Specifically, there are a lot of phenomena about non-literal language comprehension in daily life, such as jokes, sarcasm, and indirect speech, successful LLMs should be capable of applying social inference skills to attain the intended meaning beyond the literal content.
In this paper, we incorporate two kinds of social modeling into \texttt{FAC$^2$E}, encompassing pragmatics and theory of mind (ToM) reasoning.
Pragmatics is evaluated by six kinds of dialogue, including polite deceits, irony, maxims of conversation, metaphor, indirect speech, and humor, while ToM is based on the ``unexpected tasks'' devised by \citet{kosinski2023theory}.
% \citet{ullman2023large}
% which present a ToM background context in natural language, and compare the tokens generated by \emph{content prompt} and \emph{belief prompt}.
% \emph{Please refer to Appendix~\ref{sec:evaluation-data-examples} for sampled instances of each capability.}

\begin{figure*}[!t]
    \includegraphics[width=\textwidth]{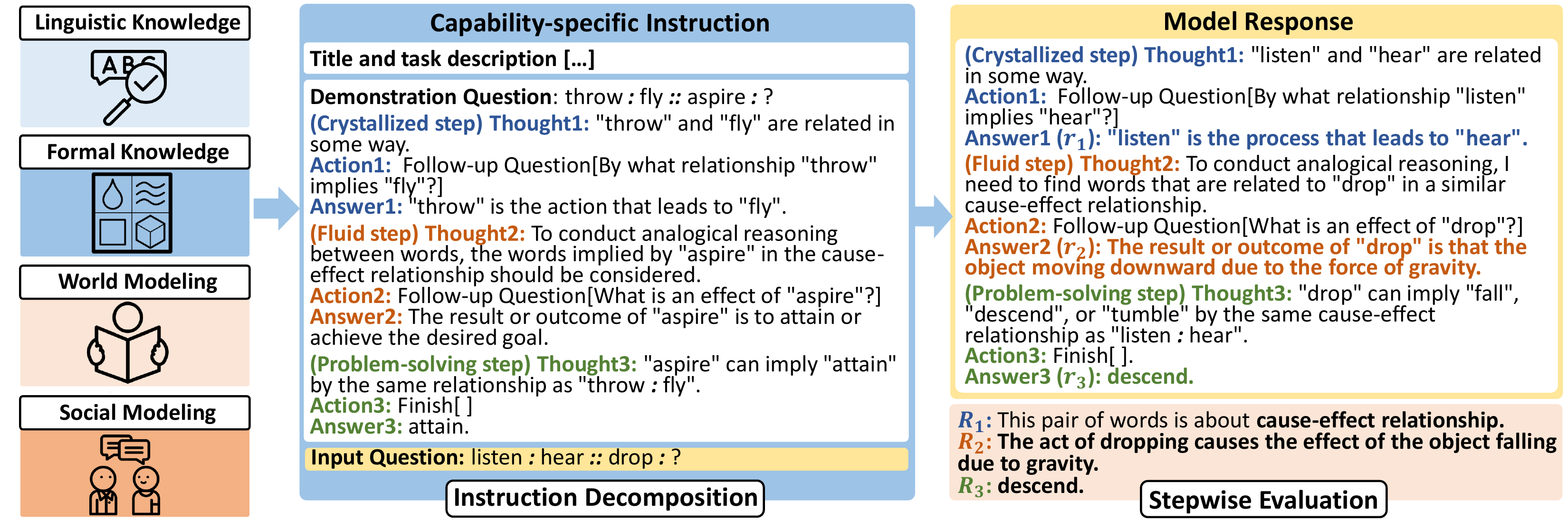}
    \caption{ Illustration of \texttt{FAC$^2$E} pipeline.
    %%
    %Based on the capability-specific instruction, 
    The input question is decomposed into two intermediate follow-up questions, which are used to help the model talk with itself to elicit reasoning sub-steps.
    \texttt{FAC$^2$E} evaluates each sub-step to reveal crystallized performance, fluid performance, and corresponding problem-solving performance. 
    The content in the round parentheses is purely illustrative and is not part of the model input.
    The instruction has been omitted here for clarity.
    Please refer to Appendix~\ref{sec:instruction-design} for full version example.
    }
    \label{fig:pipeline}
    \vspace{-5mm}
\end{figure*}

\subsection{\texttt{FAC$^2$E}}
\label{subsec:office}
Based on the formulation of LLMs' capabilities, we collect input and output pairs from various benchmarks and modify collected instances, yielding a unified question-answering (QA) format.
% filter based on length criteria, 
%%
After that, following the widely adopted few-shot in-context learning (ICL)~\cite{brown2020language}, we devise capability-specific instruction and frame \texttt{FAC$^2$E} via instruction following~\cite{ouyang2022training}.
We further leverage chain-of-thought (CoT)~\cite{wei2022emergent,wei2022chain} style prompting to elicit two intermediate reasoning steps from the model, namely \emph{crystallized} step and \emph{fluid} step. 
%, which respectively accounts for the process of knowledge recalling (\red{whether the model knows} what knowledge to use) and knowledge utilization (how well does the model adapt learned knowledge into given instances).
%%
The terms ``\emph{crystallized}'' and ``\emph{fluid}'' are borrowed from Cattel’s theory~\cite{cattell1963theory}, a foundational building block of cognitive science about the source of intelligence.
Cattel’s theory delineates that crystallized intelligence is semantic knowledge from past experiences, fluid intelligence is the ability to navigate novel situations, and problem-solving uses both.
Therefore, we add two intermediate steps to operationalize the two kinds of mechanisms, and aim to measure how well the model recalls and applies knowledge.
Last, we compare the intermediate results with the reference answers to score the reasoning sub-steps, hence providing an assessment of the crystallized performance and fluid performance as well as problem-solving performance.

As depicted in Figure~\ref{fig:pipeline}, the pipeline \texttt{FAC$^2$E} can be divided into three steps, including capability-specific instruction design, instruction decomposition, and stepwise evaluation.
Specifically, we devise natural instruction for each capability-related task.
Borrowing the widely used template schema of instruction-following~\cite{wang-etal-2022-super, mishra-etal-2022-cross}, the capability-specific instruction $\mathcal{I}_c$ is comprised of three parts: title, task description and few-shot demonstrations.
Precisely, the title defines a given QA task in high-level natural language and highlights the associated skills, while the task description not only presents a complete clarification of how an input text is expected to be mapped to final output, but also define the output of reasoning sub-steps through instruction decomposition. 
After that, following the given task description, a few in-context demonstrations are provided to better steer the response generation.
At last, we collect the response results for each reasoning sub-steps, denoted as $\{ r_i \}_{i=1}^{3}$, and respectively evaluate them with the reference answer, denoted as $\{ R_i \}_{i=1}^{3}$, which are directly extracted from corresponding benchmarks.
Formally, the procedure can be represented as:
\begin{align}
    \{ r_i \}_{i=1}^{3} &= \mathcal{M}\big( \mathcal{I}_c \big) \\
    s_i &= \text{Criterion}_i \big( r_i, R_i\big)
\end{align}
where $\mathcal{M}$ denotes the examined LLM, while $\text{Criterion}_i$ and $s_i$ represent the employed automatic metric and corresponding score, respectively.

\noindent
\textbf{Instruction decomposition.}\
% Single performance metric on given tasks can manifest little about why the model answers the given question correctly or not, rendering the lack of interpretability of evaluation results and the inconvenience of conducting fine-grained analysis.
%%
% Motivated by building explainable evaluation methods for machine reading comprehension (MRC)~\cite{ray-choudhury-etal-2022-machine}, which extracts relevant text spans and validates reasoning path, we propose to evaluate intermediate steps when LLMs apply a specific capability to solve practical problems. 
% %%
% Specifically, we decompose the capability-specific instruction and add two intermediate steps, including crystallized and fluid steps.
% %%
% In practice, we leverage CoT-like iterative prompting strategy to elicit the intermediate reasoning from the model.
We leverage CoT-like iterative prompting strategy to elicit the intermediate reasoning from the model to frame crystallized and fluid steps.
Differing from the standard CoT that outputs a continuous rationale before the final answer, we first decompose the given question as follow-up sub-questions.
After that, these sub-questions are used to help the model talk with itself to respectively discover (i) what knowledge this question is about, (ii) how to apply relevant knowledge to the given instance, and (iii) the final answer.
In other words, \texttt{FAC$^2$E} convert the CoT continuous rationale into easily parseable multi-step rationales, which externalizes reasoning of the model~\cite{shwartz-etal-2020-unsupervised, zhou-etal-2022-think} and enables the evaluation of crystallized performance and fluid performance.
%%
% Therefore, the instruction decomposition of \texttt{FAC$^2$E} can also be regarded as a kind of knowledge externalization method through inquiring LLM~\cite{shwartz-etal-2020-unsupervised, zhou-etal-2022-think, zhou2022least} with pre-defined instruction.
%%
Formally, as depicted in Figure~\ref{fig:pipeline}, we expect that the model outputs as:
% \begin{align*}
%     &[\text{\texttt{Thought i}}] \;\; [\cdots] \\
%     &[\text{\texttt{Action i}}] \;\;\;\, [\cdots] \\
%     &[\text{\texttt{Answer i}}] \;\;\;\, [\inlinetcboxmathsubscript[boxsep=3pt]{r}{i}]
% \end{align*}
$
[\text{\texttt{Thought}}], [\text{\texttt{Action}}], [\text{\texttt{Answer}}]
$
, where [\texttt{Thought}] can reason about the current situation, [\texttt{Action}] can be either (1) [\texttt{Follow-up Question}], which returns a sub-question, or (2) [\texttt{Finish}], and [\texttt{Answer}] is extracted as the reasoning result of a sub-step.

\begin{table}[!t]
    \centering
    \resizebox{1.0\columnwidth}{!}{
        \begin{tabular}{ccccc}
            \toprule
            \textbf{Capability}& \textbf{Benchmark}& \textbf{\#Samples}& \textbf{Length}& \textbf{QA} \\
            \midrule
            \midrule
            Agreements& BLiMP~\cite{warstadt-etal-2020-blimp-benchmark}& 400& 33& M \\
            % Subject-Verb& \citet{wilcox-etal-2019-structural}& xxx& M \\
            Licensing& \citet{marvin-linzen-2018-targeted}& 1,500& 36& M \\
            Long-distance dependency& \citet{wilcox-etal-2019-structural}& 660& 48&  M \\
            Garden-path effects& \citet{futrell2018rnns}& 450& 40&  M \\
            Lexical semantics& \citet{petersen-potts-2023-lexical}& 1,000& 8&  M \\
            \hline
            Deductive& \citet{bang-etal-2023-multitask}& 1,600&	18& M \\
            Inductive& \citet{bang-etal-2023-multitask}& 1,500&	16& M \\
            Analogical& \citet{webb2023emergent}& 800& 4& G \\
            Numeric& MAWPS~\cite{koncel-kedziorski-etal-2016-mawps}& 400&	33& G \\
            % Listing& 1024& 9.68& M \\
            Logic& \citet{tian-etal-2021-diagnosing}& 1,000& 87& G \\
            Manipulatation& \citet{wei2022chain}& 250& 31& G \\
            \hline
            Factual Knowledge& LAMA~\cite{petroni-etal-2019-language}& 500& 48& G \\
            Reading Comprehension& \citet{dua-etal-2019-drop}& 1,000& 195&  M \\
            Commonsense& \citet{talmor-etal-2019-commonsenseqa}& 800& 13& M \\
            Discourse& \citet{wang2023disco}& 400& 439& G \\
            Narrative& \citet{xu-etal-2022-fantastic}& 200& 395&  G \\
            % Event& 986& 9.42& M \\
            \hline
            Pragmatics& \citet{hu-etal-2023-fine}& 150& 288& M \\
            Theory of mind& \citet{ullman2023large}& 80& 152& G \\
            \bottomrule
        \end{tabular}
    }
    \caption{Breakdown statistics on source benchmarks employed by \texttt{FAC$^2$E} and re-formulation types (\textbf{\underline{G}}enerative or \textbf{\underline{M}}ultiple-choice QA), where Length refers to the average input length of examples.}
    \label{tab:data-construction}
    \vspace{-5mm}
\end{table}

\noindent
\textbf{Stepwise evaluation.} \
Given the reasoning results of three sub-steps, \ie $\{ r_i \}_{i=1}^{3}$, we engage automatic metrics as the criterion to evaluate them.
Specifically, $r_1$ and $r_2$ are free-form rationales for intermediate reasoning steps. 
Considering the diversity of rationale generation, we resort to BARTScore-Recall~\cite{yuan2021bartscore}, one of the most superior metrics for natural language generation to evaluate the quality of generated rationale automatically.
BARTScore-Recall gauges how many semantic content units from reference texts are covered by the generated candidates, and will not penalize the redundant and instance-specific information in the model response.
For the last response $r_3$, since it is expected to be the final answer for the given question, it is evaluated by the BARTScore-Recall~\cite{yuan2021bartscore} or accuracy for generative QA re-formulation and multiple choice QA re-formulation, respectively.

\section{Experiments}
\label{sec:experiments}

\noindent
\textbf{Evaluation data construction.} \
In Table~\ref{tab:data-construction}, we present a collection of 17 widely adopted English benchmarks and modify the corresponding input-output into a unified QA format, \ie generative QA or multiple-choice QA.
The choice to utilize these benchmarks is rooted in considerations of data quality, availability, and specificity of focus; hence, some widely recognized benchmarks may not be included for these reasons.
For example, PIQA~\cite{bisk2020piqa} focuses on physical commonsense, which, while valuable, represents only a single facet of commonsense reasoning. In contrast, MMLU~\cite{hendrycks2020measuring} encompasses a broad spectrum of subjects, but requires both commonsense and contextual understanding, which might not align with our goal of ensuring a broad range of capabilities while minimizing the coupling between abilities during evaluation.

The reference answers of the benchmarks are directly used as the final answer $R_3$, while the reference rationales ($R_1$ and $R_2$) for the intermediate reasoning steps are constructed automatically.
Specifically, on the one hand, $R_1$, \ie the reference rationale for the first reasoning step, is based on the rationale templates and the gold labels of the employed benchmarks.
For example, when evaluating the grammaticality regarding negative polarity item (NPI) licensing, the rationale template for $R_1$ could be ``\emph{The word $[ \; ]$ is a negative polarity item: it can only be used in the scope of negation.}''.
The blank is then filled with gold labels (licensing contexts or trigger words), such as ``\emph{any}'', ``\emph{ever}'', and ``\emph{even}'', to build the final $R_1$ for corresponding NPI licensing samples.
On the other hand, $R_2$, \ie the reference rationale for the second reasoning step is built on the instance-wise annotations of human evaluation publicly released by the authors of corresponding benchmarks, which annotates necessary explanations as well as final answer $R_3$ for a given question.
Although this will leave few benchmarks available and lead to a limited number of evaluation data, it provides relatively reliable references and especially enables reproducible evaluation.

\begin{table}[!t]
    \centering
    \resizebox{1.0\columnwidth}{!}{
        \begin{tabular}{ccccc}
            \toprule
            \textbf{Model}& \textbf{Model size}& \textbf{Pre-training}& \textbf{Fine-tuning} \\
            \midrule
            \midrule
            T5& 11B& 1.0T tokens& \XSolidBrush \\
            Flan-T5&  11B& as above& IT \\
            Flan-Alpaca&  11B& as above& IT \\
            \hline
            LLaMA& 7B& 1.4T tokens& \XSolidBrush \\
            Alpaca& 7B& as above& IT \\
            Vicuna& 7B& as above& IT \\
            TÜLU 1& 7B,13B,30B,65B& as above& IT \\
            \hline
            LLaMA 2& 7B& 2.0T tokens& \XSolidBrush \\
            LLaMA 2-Chat& 7B& as above& IT+RLHF \\
            \hline
            LLaMA 3-Instruct& 8B& 15.0T& IT+RLHF \\
            LLaMA 3.1-Instruct& 8B& 16.4T& IT+RLHF \\
            \hline
            GPT-3.5& 175B& -& IT+RLHF \\
            InstructGPT& 175B& -& IT+RLHF \\
            GPT-4& -& -& IT+RLHF \\
            Bard& 137B& -& IT+RLHF \\
            \bottomrule
        \end{tabular}
    }
    \caption{Statistics of examined LLMs, where fine-tuning techniques indicating whether the model is built with instruction tuning (IT) and reinforcement learning with human feedback (RLHF) or not.}
    \label{tab:all-models}
    \vspace{-5mm}
\end{table}

\noindent
\textbf{Examined models.} \
% We evaluate LLMs with varying model sizes, training pipelines, and training datasets. 
%%
As summarized in Table~\ref{tab:all-models}, the examined LLMs can be categorized into publicly available open-source models and proprietary ones whose responses are provided through private APIs.
Open-source models include three backbone models, \ie T5~\cite{raffel2020exploring}, LLaMA~\cite{touvron2023llama1} and LLaMA 2~\cite{touvron2023llama2}, which are pre-trained on large scale corpus and not applied to any fine-tuning.
Initialized with T5, Flan-T5~\cite{longpre2023flan} and Flan-Alpaca~\cite{chia2023instructeval} are instruction-tuned on Flan V2~\cite{longpre2023flan} and Alpaca~\cite{alpaca}, respectively.
Built on LLaMA, Alpaca~\cite{alpaca} and Vicuna~\cite{chiang2023vicuna} are instruction-tuned with responses generated by GPT-3.5, while TÜLU 1~\cite{wang2023far} are instruction-tuned with a mixture of both manually curated and distilled dataset.
Based on LLaMA 2, LLaMA 2-Chat~\cite{touvron2023llama2} is firstly instruction-tuned with high-quality collected annotations, and then aligned with human preferences for the chat use case.
LLaMA 3-Instruct and LLaMA 3.1-Instruct respectively fine-tune LLaMA 3 and LLaMA 3.1 to better understand and follow human instructions.
Besides, to perform a fair comparison w.r.t instruction-tuning dataset, 
we also evaluate LLaMA checkpoints fine-tuned on other datasets, such as Flan V2~\cite{longpre2023flan} (human-written), Alpaca (model-generated)~\cite{alpaca}, ShareGPT (user prompt with model response).

Proprietary models consist of OpenAI's GPT-3.5 (\texttt{gpt-3.5-turbo})~\cite{chatgpt}, InstructGPT (\texttt{gpt-3.5-turbo-instruct})~\cite{ouyang2022training}, GPT-4 (\texttt{gpt-4-turbo})~\cite{achiam2023gpt}, and Google's Bard~\cite{bard} (also known as Gemini~\cite{team2023gemini}).
%%
% \emph{Please refer to Appendix~\ref{sec:implemetation-details} for checkpoint details.}

\begin{table*}[!t]
    \centering
     \resizebox{1.0\textwidth}{!}{
        \begin{tabular}{c|ccc|ccc|ccc|ccc}
            \toprule
            \multirow{2}{*}{\textbf{Model}}& \multicolumn{3}{C{5cm}|}{\textbf{\textsc{Linguistic Knowledge}}}& \multicolumn{3}{C{5cm}|}{\textbf{\textsc{Formal Knowledge}}}& \multicolumn{3}{C{5cm}|}{\textbf{\textsc{World Modeling}}}& \multicolumn{3}{C{5cm}}{\textbf{\textsc{Social Modeling}}} \\
            & \multicolumn{1}{C{1.4cm}}{$s_1$}& \multicolumn{1}{C{1.4cm}}{$s_2$}& \multicolumn{1}{C{1.4cm}|}{$s_3$}& \multicolumn{1}{C{1.4cm}}{$s_1$}& \multicolumn{1}{C{1.4cm}}{$s_2$}& \multicolumn{1}{C{1.4cm}|}{$s_3$}& \multicolumn{1}{C{1.4cm}}{$s_1$}& \multicolumn{1}{C{1.4cm}}{$s_2$}& \multicolumn{1}{C{1.4cm}|}{$s_3$}& \multicolumn{1}{C{1.4cm}}{$s_1$}& \multicolumn{1}{C{1.4cm}}{$s_2$}& \multicolumn{1}{C{1.4cm}}{$s_3$}  \\
            \midrule
            \midrule
            T5& 83.99& 26.39& 47.39& 77.97& 28.10& 33.26& 74.61& 24.74& 26.53& 66.79& 18.31& 19.16 \\
            Flan-T5& 84.96& 42.50& 64.12& 80.10& 35.22& 42.58& 74.82& 36.13& 34.64& 67.73& 27.23& 21.27 \\
            Flan-Alpaca& 85.25& 39.41& 60.15& 79.88& 34.99& 41.72& 75.54& 37.42& 36.57& 68.38& 28.74& 23.82 \\
            \hline
            LLaMA& 85.34& 31.36& 53.77& 80.02& 31.18& 40.04& 75.57& 27.46& 30.47& 67.54& 20.14& 20.75 \\
            Alpaca& 86.02& 45.78& 68.39& 82.03& 38.10& 53.85& 77.30& 41.91& \centeronewlight{49.42}& 69.93& 29.61& \centeronewdeepnobar{32.96} \\
            Vicuna& 85.23& 47.45& 72.66& 84.35& 40.10& 57.07& 75.33& 43.38& 44.37& 65.68& 26.87& 30.63 \\
            TÜLU 1& 84.14& 45.84& 70.72& 82.32& 39.29& 51.21& 75.90& 43.30& 40.80& 69.73& 26.77& 27.02 \\
            \hline
            LLaMA 2& 83.19& 34.56& 57.89& 82.19& 34.18& 46.15& 77.48& 34.84& 40.92& 68.22& 24.74& 24.20 \\
            LLaMA 2-Chat& 87.04& 48.95& 74.46& 84.05& 43.21& \centeronewlight{57.13}& 78.43& 46.09& 44.46& 71.06& 28.89& 29.59 \\
            \hline
            LLaMA 3-Instruct&	87.78&	50.20&	\centeronewlight{78.57}&	85.81&	43.88&	60.79&	80.18&	46.74&	47.72&	76.99&	39.10&	30.52 \\
            LLaMA 3.1-Instruct&	88.21&	51.47&	\centeronewdeep{82.71}&	87.33&	44.94&	\centeronewdeep{65.01}&	81.54&	47.11&	\centeronewdeep{50.39}&	77.78&	30.51&	\centeronewlightnobar{31.75} \\
            \hline
            GPT-3.5& 87.91& 53.91& 82.72& 85.93& 45.20& 70.47& 81.53& 53.18& \centeronerlight{67.68}& 77.23& 36.34& 40.56 \\
            InstructGPT& 88.52& 55.50& 85.19& 85.12& 44.18& 67.48& 80.34& 51.78& 65.16& 74.17& 39.90& \centeronerdeepnobar{45.95} \\
            GPT-4& 89.32& 58.98& \centeronerdeep{89.62}& 87.64& 47.99& \centeronerdeep{75.97}& 81.86& 54.78& \centeronerdeep{69.43}& 81.24& 40.99& \centeronerlightnobar{45.71} \\
            Bard& 87.74& 52.37& \centeronerlight{86.16}& 86.97& 46.08& \centeronerlight{71.62}& 79.30& 49.09& 61.31& 78.64& 38.27& 42.53 \\
            
            \bottomrule
        \end{tabular}
     }
    \caption{Quantitative results in terms of four capability dimensions.
    As stated in Section~\ref{subsec:office}, $s_1$, $s_2$, and $s_3$ refer to crystallized performance, fluid performance, and problem-solving performance, respectively.
    %%
    % The shade of the \colorbox{r1}{red} and \colorbox{w1}{blue} text highlights the ranking of $s_3$ in open-source and proprietary models, respectively; for example, the darker shade represents the highest score, while the lighter shade represents the second highest score.
    The color of the text indicates the model type: \colorbox{w1}{blue} for open-source and \colorbox{r1}{red} for proprietary models. The shade represents the ranking, where the darker shade represents the highest score, and the lighter shade represents the second highest score.
    }
    \vspace{-5mm}
    \label{tab:main-results}
\end{table*}

\subsection{Main results}
\label{subsec:main-results}

\begin{figure}[!t]
\centering
    \includegraphics[width=1.0\columnwidth]{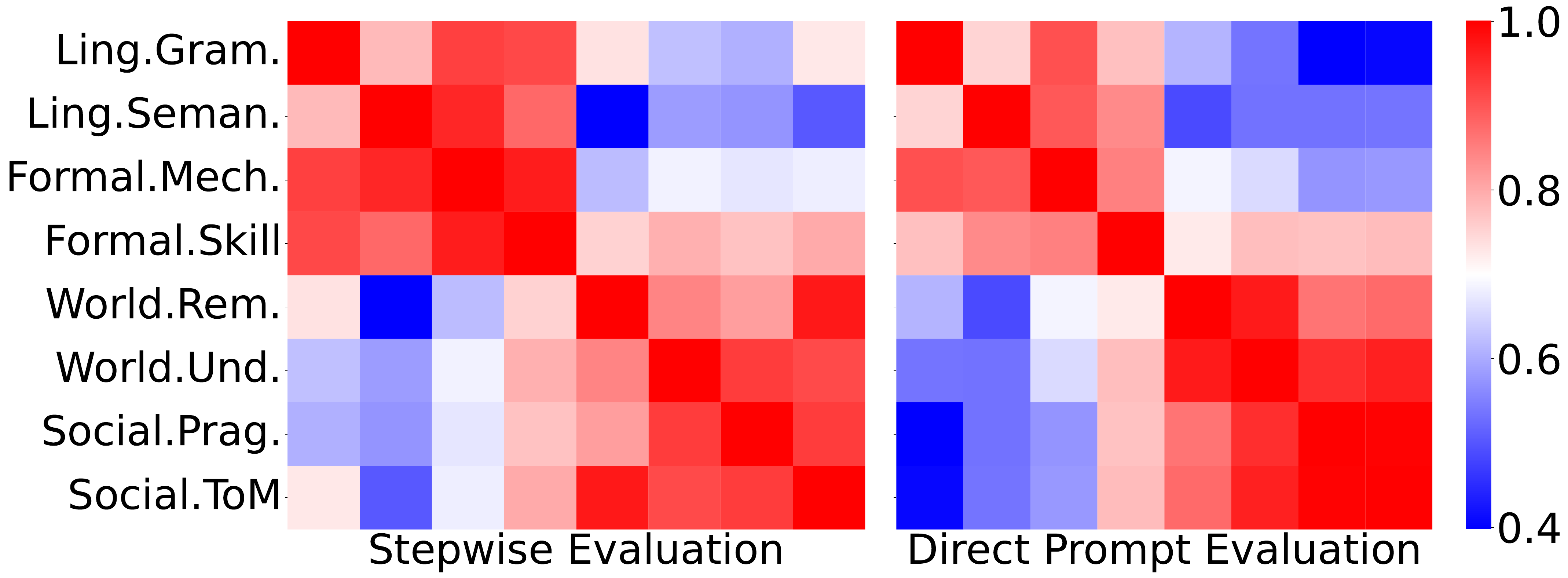}
    \caption{Pairwise correlation of problem-solving performance ($s_3$) among different capabilities.
    Please refer to Table~\ref{tab:capability-schema} for full label names.
    }
    \label{fig:skill-correlation}
    \vspace{-5mm}
\end{figure}

\noindent
\textbf{The difference in problem-solving performance is significantly greater than the difference in crystallized performance.}\
On the one hand, in terms of problem-solving performance ($s_3$), as shown in Table~\ref{tab:main-results}, open-source models usually underperform the proprietary ones across various capabilities, especially cognition-related ones, such as world modeling and social modeling.
For example, the most competitive open-source model LLaMA 3.1-Instruct achieves a 50.39 accuracy in the world modeling dimension, while GPT-4 produces a performance of 69.43, excelling Alpaca by a substantial margin (around 40\%).
A similar conclusion can also be drawn from the other dimensions, such as the best problem-solving performance of proprietary models exceeds that of open-source models by about 20\%, 32\%,  and 40\% in linguistic knowledge, formal knowledge, and social modeling, respectively.
%%
% As for the language-related capability, although \texttt{FAC$^2$E} incorporate the recently proposed challenging benchmarks, including the grammatical phenomena such as intricate long-distance dependencies, both open-source and proprietary perform well on them.
% %%
% This finding is consistent with previous single benchmark-based results~\cite{}, \ie LLMs have mastered formal linguistic competence.
On the other hand, in terms of crystallized performance ($s_1$), there is a rather smaller gap between open-source and proprietary models compared to problem-solving accuracy.
For example, the maximal difference of $s_1$ in the world modeling among all the examined models is about 9\%, \ie GPT-4's 81.86 vs. T5's 74.61, while their difference of $s_3$ is GPT-4's 69.43 vs. T5's 26.53.
This inconsistency between $s_1$ and $s_3$ can also be observed in other capability dimensions, potentially showing that either pre-training or fine-tuning of LLMs can encode sufficient knowledge into the model, but the final task performance does not just depend on the amount or quality of knowledge.

Notably, the crystallized performance (s1) of LLaMA 3-Instruct and LLaMA 3.1-Instruct has significantly improved compared to LLaMA 2-Chat. For example, in social modeling, LLaMA 3-Instruct scored 76.99, and LLaMA 3.1-Instruct scored 77.78, compared to LLaMA 2-Chat's 71.06. This suggests that these models better encode knowledge relevant to higher-level cognitive tasks, likely due to training and fine-tuning LLaMA 3 on higher-quality and larger-scale data compared to the LLaMA 2 model.

\begin{figure}[!t]
\centering
    \includegraphics[width=1.0\columnwidth]{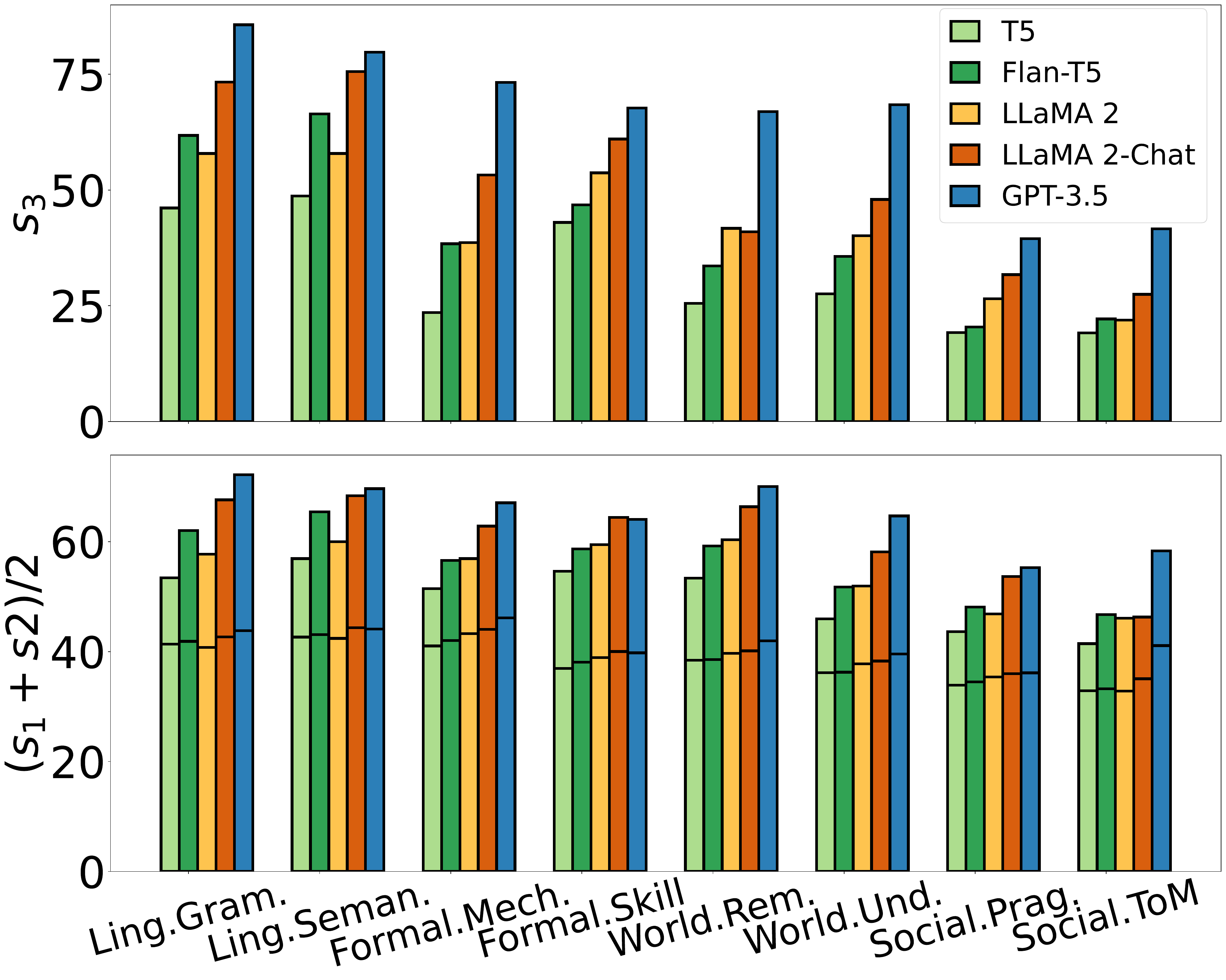}
    \caption{Bar diagram illustrating the relationship between problem-solving performance ($s_3$) and intermediate performance ($(s_1 + s_2) / 2$).
    Each bar of intermediate performance is divided into two stacked segments, the \emph{lower} one denotes $s_1$, while the \emph{upper} one denotes $s_2$.
    }
    \label{fig:decomposed-performance}
    \vspace{-5mm}
\end{figure}

\noindent
\textbf{Linguistic capabilities show a relatively weak correlation with cognitive capabilities.} \
Figure~\ref{fig:skill-correlation} presents the correlation results between different capabilities.
Both the language-related and cognition-related capabilities exhibit stronger (Pearson's $r$ > 0.7~\cite{krippendorff2004reliability}) intra-dimension correlation (\eg world modeling vs. social modeling) when compared to inter-dimension correlation (\eg world modeling vs. linguistic knowledge).
This indicates that excellence in language processing does not necessarily equate to a similar level of cognitive capability.
%%
% This result can also be observed from Table~\ref{tab:main-results}.
% %%
% For example, the model LLaMA 2-Chat achieves better results in terms of problem-solving performance of linguistics knowledge than Alpaca, \ie LLaMA 2-Chat's 74.46 vs. Alpaca's 68.39, but it fails to surpass Alpaca in world modeling (44.46 vs. 49.42) and social modeling (29.59 vs. 32.96).
% %%
% This pattern also applies to other models, such as Bard vs. InstructGPT, further demonstrating the reasonability of disassociating language-related capabilities and cognition-related ones to better evaluate LLMs.
%%
These results can also be observed from direct prompt evaluation, which assesses problem-solving performance ($s_3$) without prompting intermediate reasoning steps, further demonstrating the rationality of dissociating language and cognition.

A possible reason behind it could be that dedicated structures of the model or subsets of parameters are highly correlated with language~\cite{zhang2024unveiling, tang2024language}, whereas others serve as cognition~\cite{chen2023journey}, and they are optimized at different training stages and function as different mechanisms during inference, which has been verified by recent studies on knowledge locating and editing of LLMs~\citet{dai-etal-2022-knowledge, meng2022mass, zhang2024comprehensive}.
%%
% This result can also be regarded as an empirical explanation for recent studies on knowledge locating and editing of LLMs. 
%%
% For example, \citet{zhang2024unveiling} and \citet{chen2023journey} respectively found core linguistic regions and language-independent knowledge neurons in LLMs, and \citet{dai-etal-2022-knowledge, meng2022mass} proposed to explicitly edit knowledge neurons to improve performance, potentially providing a feasible approach to strengthening LLMs capability without fine-tuning, \ie editing their parameters or architectures directly.

\begin{figure}[!t]
\centering
    \includegraphics[width=0.8\columnwidth]{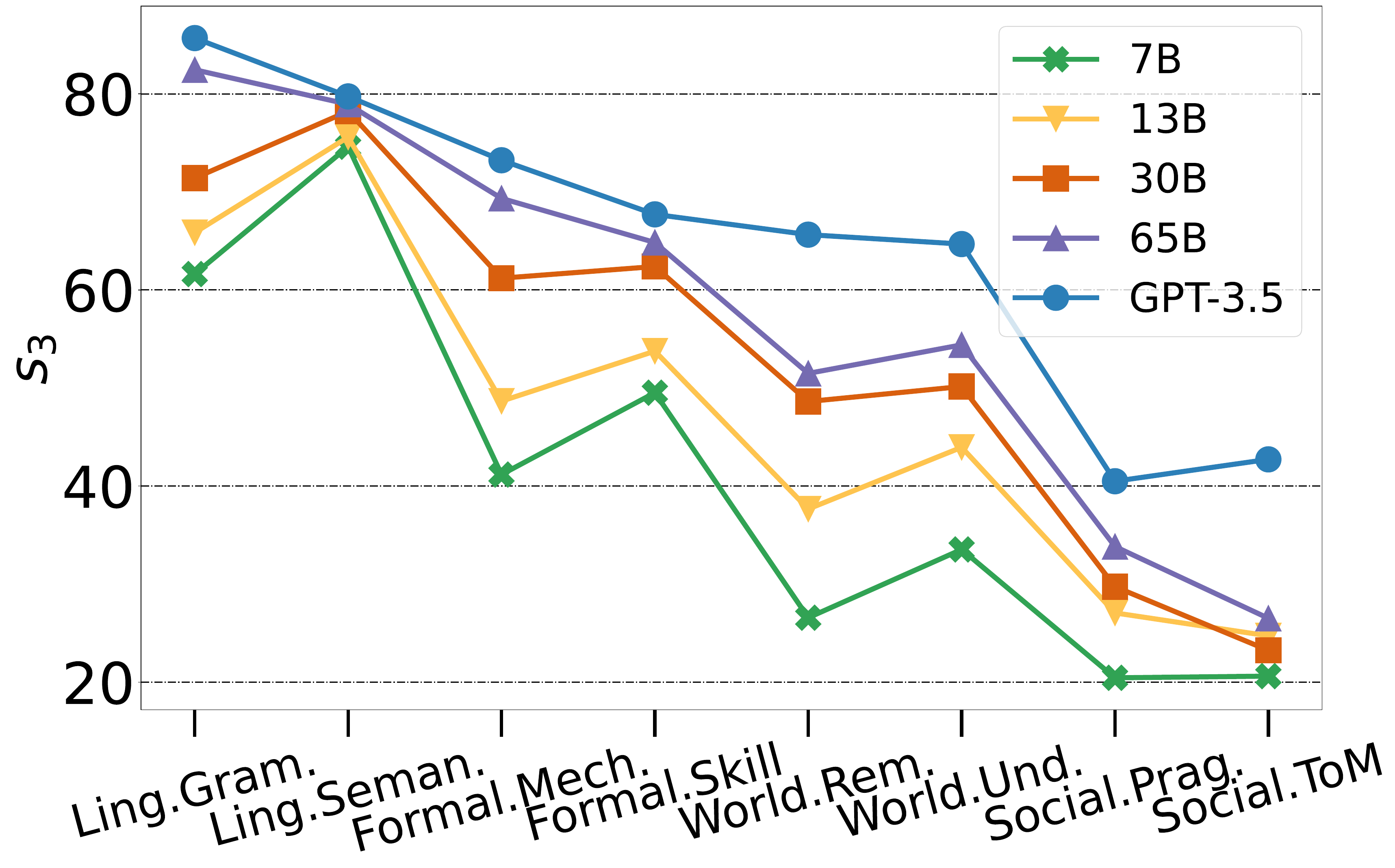}
    \caption{
    Problem-solving performance of instruction-tuned LLaMA with different model sizes.
    }
    \label{fig:different-model-size}
    \vspace{-2mm}
\end{figure}

\begin{figure}[!t]
\centering
    \includegraphics[width=0.8\columnwidth]{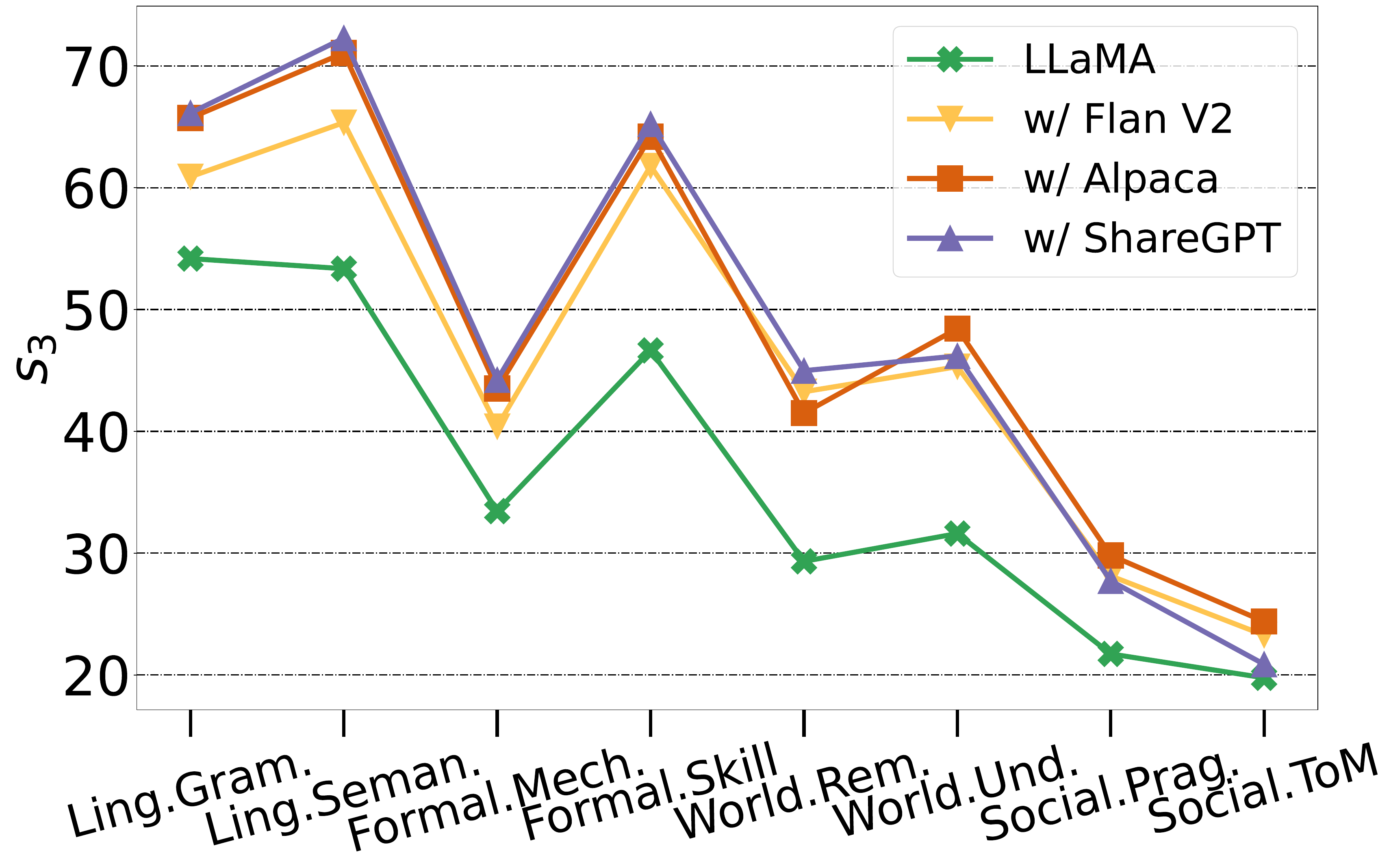}
    \caption{
    Problem-solving performance of LLaMA on different instruction-tuning datasets.
    }
    \label{fig:different-instruction-dataset}
    \vspace{-5mm}
\end{figure}

\noindent
\textbf{The crystallized step impacts problem-solving more than the fluid step.} \
Figure~\ref{fig:decomposed-performance} illustrates the relationship between problem-solving performance ($s_3$) and sum of crystallized ($s_1$) and fluid ($s_2$) performance.
Both $s_1$ and $s_2$ make a difference to the final $s_3$, showing that problem-solving not only depends on the amount or quality of stored knowledge but also is reflective of the effectiveness of knowledge utilization.
For example, LLaMA 2 underperforms LLaMA 2-Chat in terms of $s_3$ across various capabilities.
%% such as world understanding and social pragmatics.
%%
When taking a closer look at the intermediate results, we can observe that both the models do well in crystallized step, but LLaMA 2 shows an worse result in fluid step, leading the worse problem-solving performance than LLaMA 2-Chat.
Besides, all of the open-source models exhibit relatively poor fluid performance w.r.t. GPT-3.5, especially those that are pre-trained but not instruction-tuned, such as T5 and LLaMA 2.
%%
% issues on knowledge utilization and instruction-oriented full-model fine-tuning?
%%
% This not only provides an auxiliary explanation of what the model has learned from the instruction-tuning dataset, \ie focusing on knowledge utilization by following instructions~\cite{wei2022emergent}, but also 
This implies a solution to improve problem-solving performance, \ie boosting the efficacy of knowledge utilization.
Section~\ref{subsec:booting-llms} presents a knowledge-enhanced method to demonstrate this solution.

\noindent
\textbf{Both the model size and the quality of fine-tuning dataset affect the capabilities of LLMs.} \
Firstly, backbone models play a critical role in building superior models.
For example, as shown in Table~\ref{tab:main-results}, fine-tuned on the same dataset, LLaMA-based Alpaca performs better than T5-based Flan-Alpaca, and the larger scale proprietary models show a greater advantage over other open-source models.
In addition, scaling open-source models does improve both language and cognitive capability.
As shown in Figure~\ref{fig:different-model-size}, 
problem-solving performance across various capabilities increases as the model size increases, and 65B achieves the best performance. 
In particular, the level of formal knowledge of 65B is close to that of GPT-3.5.
Last, but not least, there is no significant performance difference among various open-source instruction-tuning datasets whether it is comprised of human-written instruction or not.
As illustrated in Figure~\ref{fig:different-instruction-dataset}, there is not a single best instruction tuning dataset across all tasks, indicating different datasets bring different benefits to LLMs' capabilities.
This finding is consistent with the recent success of a mixture of instruction-tuning datasets or expert LLMs~\cite{jiang2024mixtral, xia2024less}.
%, which exhibits great traction to overall capability.

% \begin{figure}[!t]
%     \centering
% 	\begin{subfigure}{0.32\columnwidth}
%         \includegraphics[width=\columnwidth]{figures/RAG1.png}
%         \subcaption{}
%         \label{fig:no-knowledge}
% 	\end{subfigure}
%         \hfill
% 	\begin{subfigure}{0.32\columnwidth}
%         \includegraphics[width=\columnwidth]{figures/RAG2.png}
%         \subcaption{}
%         \label{fig:r1-knowledge}
%         \end{subfigure}
%         \hfill
% 	\begin{subfigure}{0.32\columnwidth}
%         \includegraphics[width=\columnwidth]{figures/RAG3.png}
%         \subcaption{}
%         \label{fig:r2-knowledge}
%         \end{subfigure}
%     \caption{Comparisons of pipelines between knowledge-augmented baselines (b) $\mathcal{M}$+$R_1$ and (c) $\mathcal{M}$+$R_1$+$R_2$ and original setting (a).}
%     \label{fig:rag-baselines}
% \end{figure}
\begin{figure}[!t]
    \centering
    \includegraphics[width=\columnwidth]{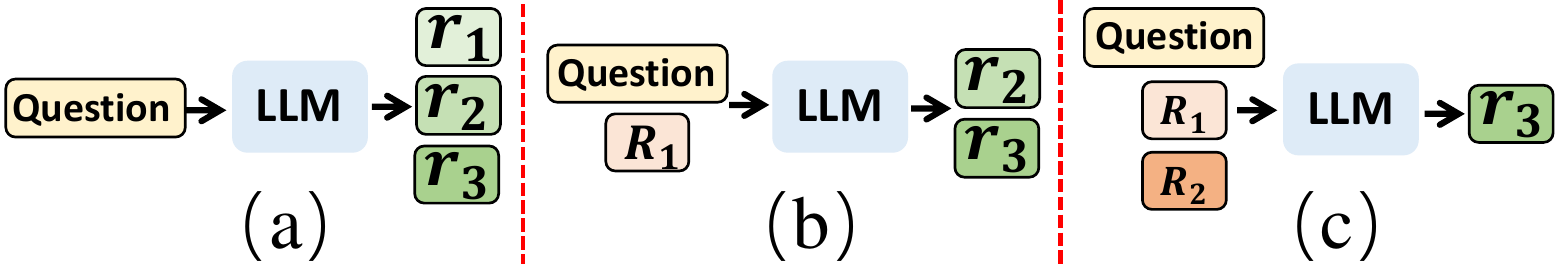}
    \caption{Comparing knowledge-enhanced baselines $\mathcal{M}$+$R_1$ (b), $\mathcal{M}$+$R_1$+$R_2$ (c) and the original setting (a).}
    \label{fig:rag-baselines}
    \vspace{-5mm}
\end{figure}

\subsection{Boosting LLMs with Injected Knowledge}
\label{subsec:booting-llms}
Based on the above analysis showcasing the limitations of the crystallized performance of existing LLMs, as illustrated in Figure~\ref{fig:rag-baselines}, we propose a knowledge-enhanced approach.
Specifically, for each given instance with a question and answer, the first baseline, denoted as $\mathcal{M}$+$R_1$, append the first reference rationale, \ie $R_1$, to the input question with string concatenation.
Then the augmented input is fed into the model with the same instruction as the examined model.
Note that we also remove the first triplet of $\langle$ [\texttt{thought}], [\texttt{action}], [\texttt{answer}] $\rangle$ in the input demonstrations for the $\mathcal{M}$+$R_1$ baseline because we have provided the corresponding reference rationale.
As a comparison, following a similar procedure, we also construct another baseline by incorporating both $R_1$ and $R_2$ into the model, denoted as $\mathcal{M}$+$R_1$+$R_2$.
%, report the results of $s_3$ for it.

\begin{figure}[!t]
    \centering
    \includegraphics[width=0.85\columnwidth]{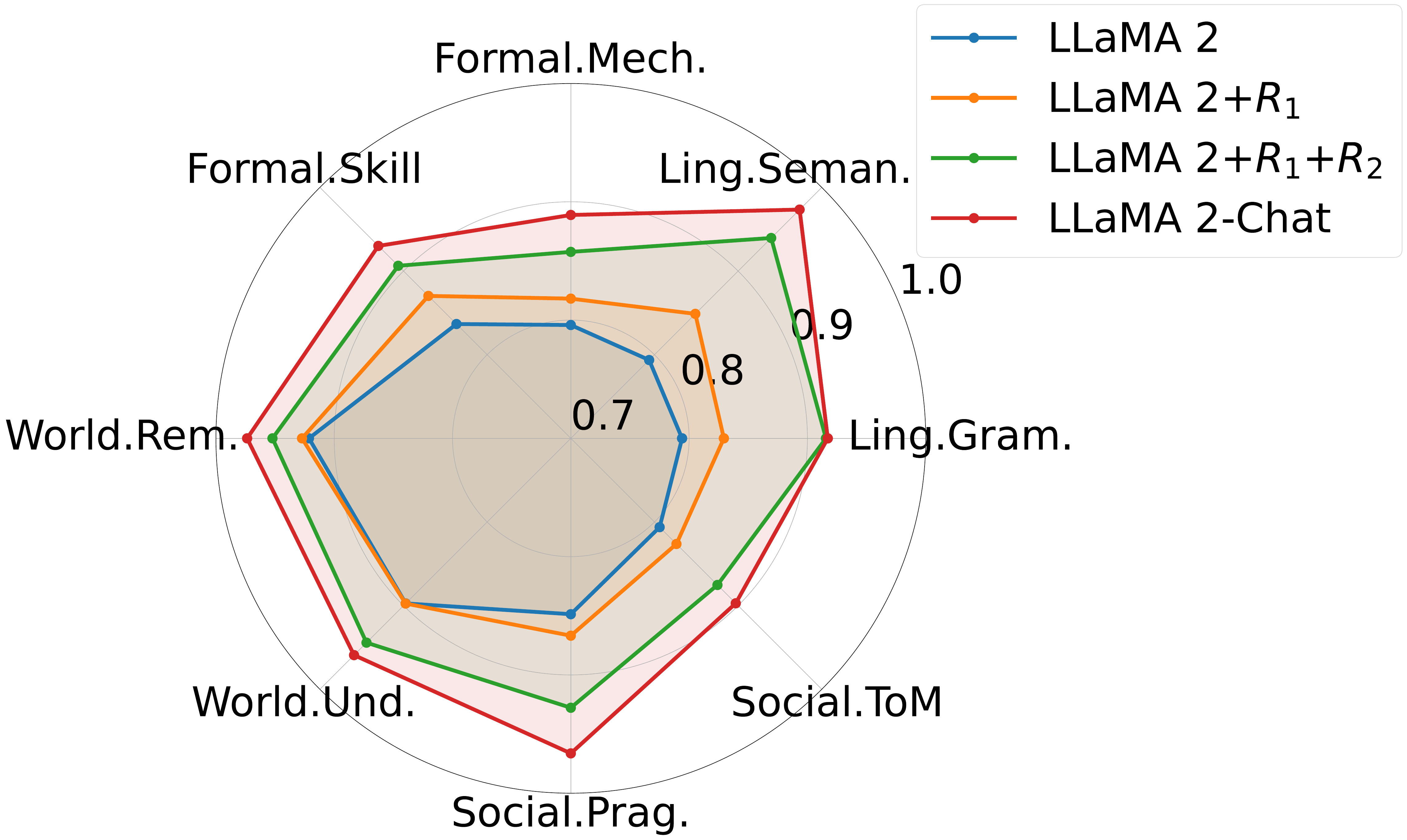}
    \caption{A 8-dimensional capability map of LLaMA 2 model when augmented with different knowledge text.
    The score is re-scaled through max-min normalization among each capability for clarity.
    }
    \label{fig:boosting-llms}
    \vspace{-5mm}
\end{figure}

Taking LLaMA 2, performing moderately in Table~\ref{tab:main-results}, as the backbone model, the multifaceted results are summarized in Figure~\ref{fig:boosting-llms}. 
We can observe that explicit injected rationales both $R_1$ and $R_2$ can substantially improve the problem-solving performance, and $R_2$ results in more improvements than $R_1$.
Specifically, $R_1$ contributes slightly to language-related capabilities, such as linguistic semantics and formal skills, while $R_2$ brings about significant improvements to cognition-related ones, especially social modeling.
%%
% This result further justifies the difference between crystallized and fluid mechanisms and function of instruction-tuning~\cite{}.
%%
Overall, the knowledge-enhanced LLaMA 2 baseline can achieve approximately 90\% performance compared to the corresponding instruction-tuned variant.
%%
% \emph{Please refer to Appendix~\ref{sec:more-results}} for more results.
% %%
% These results not only agree with the findings of retrieval-augmented generation but also raise the efficacy issue of instruction-tuning, \ie how case? and how to improve
% %%
% Besides, taking a closer look at the difference of increasing capability level, we can find that language-related vs. cognition-related.

\section{Related Works}
\label{sec:related works}

\noindent
\textbf{Evaluation of LLMs.} \
% The most prior studies evaluate LLMs on various natural language processing tasks, including both 
LLMs are initially assessed on various understanding~\cite{wang-etal-2018-glue, wang2019superglue} and generation tasks~\cite{pilault-etal-2020-extractive, thompson-post-2020-automatic}.
With the increasing emphasis on the trustworthiness of models, dedicated benchmarks are proposed to evaluate robustness~\cite{yang2022glue, wang2023robustness}, hallucination~\cite{li-etal-2023-halueval, belyi2024luna}, and
generalizability~\cite{wang2023decodingtrust}.
% bias~\cite{zhong2023agieval}
% %%
% For better domain-specific or scenario-specific applications, some formulate benchmarks to assessing natural science~\cite{arora2023have, castro2023large}, and use of external tools~\cite{huang2024planning}. 
%%
%In a more unified form,
Recent-emerged works evaluate LLMs using another evaluator LLM~\cite{kim2024prometheus, kim2024prometheus2} or on a holistic benchmark~\cite{mialon2023gaia, rein2023gpqa}.
\citet{chia2023instructeval} conduct evaluation from problem-solving, writing, and human alignments, while \citet{ye2024flask} annotates a single instance with a set of skills, including logical thinking, background knowledge, problem handling, and user alignment.
Although they provide fine-grained analysis of LLMs' capability, they suffer from the limited number of testing instances, \eg 1,700 of \citet{ye2024flask} and overlook the language-related low-level capabilities.
%%
% We consider the lessons of these works and define the LLMs' decor-related capabilities from the cognitive nature of human language and decompose the evaluation process for better interpretation.

\noindent
\textbf{Cognition-inspired intelligence evaluation.} \
How to define and evaluate intelligence is widely investigated by both cognitive science and AI benchmarking~\cite{cattell1963theory,rogers2023qa}.
In the MRC evaluation, \citet{chollet2019measure} describes intelligence as skill-acquisition efficiency,
% and highlighting the concepts of scope, generalization difficulty, priors, and experience, while \citet{sugawara2020assessing} propose to benchmark MRC based on reading and reasoning skills.
%%
while \citet{sugawara2020assessing}, \citet{wang-etal-2022-feeding} and \citet{ray-choudhury-etal-2022-machine} propose to benchmark MRC through reasoning skills and steps a system would be ``reading slowly''.
%%
% Similarly, aiming at an interpretable evaluation of MRC, 
% \citet{ray-choudhury-etal-2022-machine} draw inspirations from philosophy of mind and pedagogy, and define the reasoning steps a system would be ``reading slowly''.
%%
% As for the evaluation of LLMs, \citet{guo2023evaluating} and \citet{chang2023survey} summarize the grand challenge of existing evaluation setup, \ie not enough to thoroughly evaluate the true cognitive capabilities of LLMs.
%%
As for the LLMs, \citet{mahowald2023dissociating} summarize extensive neuroscience evidence of human language and propose a conceptual framework with formal and functional competencies, which largely motivated the design of \texttt{FAC$^2$E}. 
Compared to \citet{mahowald2023dissociating}, \texttt{FAC$^2$E} introduces more concrete NLP tasks with a more comprehensive capability system and leverages stepwise evaluation to improve the accuracy of assessments.

\section{Conclusion}
\label{sec:conclusion}
We present \texttt{FAC$^2$E}, defining a fine-grained capability evaluation framework for LLMs, which decomposes each capability into sub-steps to assess performance of knowledge recalling and utilization as well as problem-solving.
\texttt{FAC$^2$E} reveals the limitations of existing LLMs in knowledge utilization and provides a knowledge-enhanced remedy for it.
Empirical results demonstrate its effectiveness.
% facilitating better model development and benchmark construction.

\section*{Limitations}
\label{sec:limitations}

Our work proposes a fine-grained and cognition-grounded capability evaluation framework for LLMs, namely \texttt{FAC$^2$E}, which is based on the dissociated relationship of language and cognition, and evaluating the intermediate reasoning steps of LLMs.
The limitations are two-fold, including data quality and domain generalizability.

On the other hand, motivated by a variety of empirical evidence from both neuroscience and probing experiments of LLMs, we formulate \texttt{FAC$^2$E} as four capability dimensions, then re-formulate instances from multiple existing benchmarks and conduct stepwise evaluation.
Although we try to ensure that the employed datasets are as consistent and targeted with the defined dimensions as possible, the kind of evaluation data construction might not reflect the required skills accurately.
For example, an instance can cover more than one language or cognition capabilities
As discussed in Section~\ref{subsec:main-results}, different capabilities are correlated to each other to some extent, 
Besides, the reference rationales, \ie the gold standard of the intermediate reasoning step, are based on the human annotations from original benchmarks, leading to the inconsistency of reference answers and a limited number of available data, which might bias the evaluation results.
One remedy to these incidental issues could be building a new holistic benchmark with fine-grained annotations following our proposed schema. 
We regard it as our future work and deem designing a new annotation specification a promising direction.

On the other hand, our \texttt{FAC$^2$E} only examined LLMs on general domains and English input, ignoring the domain-specific and multilingual application. 
In particular, the reasoning process of LLMs may expose social bias encoded in these models, such as race and gender~\cite{lucy-bamman-2021-gender}.
Therefore, additional evaluation protocols considering potential risks to user safety are left for our future work.

\section*{Ethics Statement}
We introduce \texttt{FAC$^2$E}, a fine-grained and cognition-grounded capability evaluation framework for LLMs, and conduct evaluation experiments on publicly available datasets which are widely used in related research. Although LLMs have the potential to cause harm at the individual and societal levels~\cite{gonen-goldberg-2019-lipstick-pig}, our \texttt{FAC$^2$E} aims to provide a deep understanding of the capabilities and limitations of LLMs, potentially making the risks from the LLMs more predictable.

\section*{Acknowledgements}
This work is supported by the Canada CIFAR AI Chair Program and the Canada NSERC Discovery Grant (RGPIN-2021-03115).

% \bibliography{custom,anthology1,anthology2}
\bibliography{allinone}

\clearpage
\appendix
\section{Implementation Details}
\label{sec:implemetation-details}
All of the examined open-source models are based on HuggingFace Transformers package~\cite{wolf-etal-2020-transformers}.
Their model cards, \ie checkpoints consist of: 
\begin{itemize}
    \item T5 ({t5-11b}),
    \item Flan-T5 ({google/flan-t5-xxl}),
    \item Flan-Alpaca ({declare-lab/flan-alpaca-xxl}),
    \item LLaMA~\footnote{\url{https://huggingface.co/docs/transformers/main/en/model_doc/llama}}, 
    \item Alpaca and LLaMA on Alpaca ({allenai/open-instruct-stanford-alpaca-13b}),
    \item Vicuna ({lmsys/vicuna-7b-v1.1}),
    \item TÜLU 1 ({allenai/tulu-7b}, {allenai/tulu-13b}, {allenai/tulu-30b}, {allenai/tulu-65b}),
    \item LLaMA on Flan V2 ({allenai/open-instruct-flan-v2-13b}),
    \item LLaMA on ShareGPT ({allenai/open-instruct-sharegpt-13b}),
    \item LLaMA 2 ({meta-llama/Llama-2-7b-hf}),
    \item LLaMA 2-Chat ({meta-llama/Llama-2-7b-chat-hf})
\end{itemize}
For the response generation of each target model, as suggested by \citet{wei2022chain,zhou2022least}, we employ 4-shot instruction-following settings, \ie 4 in-context demonstrations in the input prompt, set the temperature to 0.7 and set the max length of generated sequences as 1024.
For automatic metrics, we leverage the official implementation of BARTScore~\cite{yuan2021bartscore}.

After collecting instances from various benchmarks as summarized in Table~\ref{tab:data-construction}, we remove those instances where the input length is longer than 2048, maximal context length during training except T5, Flan-T5, and Flan-Alpaca,

We conduct evaluation experiments on 2 A100 GPUs and report the average results of a total of ten runs for each model on each benchmark.
For the capabilities involving multiple benchmarks, the overall score are calculated as the arithmetic mean of crystallized performance ($s_1$), fluid performance ($s_2$), or problem-solving performance ($s_3$).

\section{Instruction Design}
\label{sec:instruction-design}
See Figure~\ref{fig:instruction-example-1} and Figure~\ref{fig:instruction-example-2} for full version example of capability-specific instruction when evaluating the analogical reasoning and grammaticality.

% \section{Evaluation Data}
% \label{sec:evaluation-data-examples}

% \noindent
% \textbf{Linguistics Knowledge} are evaluated from gramaticality and shallow semantics.
% %%
% on the one hand, grammaticality includes four kinds of skills, 
% \ie agreements, licensing, long-distance dependencies, and garden-path effects.
% and original minimal pairs tasks, where the minimal pairs consist of two sentences that differ by one or several words, one of them is grammatical, but another is ungrammatical, are re-formulated as multiple-choice QA task (two options).
% \begin{itemize}
%     \item \textbf{agreements}~\cite{}
% \end{itemize}

% \section{Case Study}
% \label{sec:case-study}

% \section{More Results}
% \label{sec:more-results}

\begin{figure*}[!t]
\begin{tcolorbox}
    \small
    Instruction: Solve a question-answering task by conducting analogical reasoning between words.
    
    Given three words, i.e. A, B, and C in a format of A:B::C:?, which means that A implies B by some relationship, reason with this relationship and predict a word D such that C implies D by the same relationship. In other words, A:B is a reference pair in some relationship, complete the pair of C:D in the same relationship as A:B.
    
    Please solve the task by interleaving Thought, Action, and Answer steps. 
    Thought can reason about the current situation, and Action can be the following two types:
    
    (1) Follow-up Question[question], which returns a sub-question with a single answer that helps solve the original question.
    
    (2) Finish[], which means no more sub-questions. The final answer should be generated in the following line.
    
    [Demonstration Question]: throw:fly::aspire:?
    
    [Thought 1]: "throw" and "fly" are related in some way.
    
    [Action 1]: Follow-up Question[By what relationship "throw" implies "fly"?]
    
    [Answer 1]: "throw" is the action that leads to flying "fly".
    
    [Thought 2]: To conduct analogical reasoning between words, the words implied by "aspire" in the cause-effect relationship should be considered.
    
    [Action 2]: Follow-up Question[What is an effect of "aspire"?]
    
    [Answer 2]: The result or outcome of "aspire" is to attain or achieve the desired goal.
    
    [Thought 3]: "aspire" can imply "attain"  by the same relationship as "throw:fly".
    
    [Action 3]: Finish[]
    
    [Answer 3]: attain.
    
    [More Demonstration Questions] [...]
    
    [Input Question]: listen:hear::drop:?
\end{tcolorbox}
\caption{Full version example of the capability-specific instruction.}
\label{fig:instruction-example-1}
\end{figure*}

\begin{figure*}[!t]
\begin{tcolorbox}
    \small
    Instruction: Solve a question-answering task judging which one of the minimal pairs is acceptable and grammatical.

    The minimal pairs consist of two sentences that differ by a few words, one of them is grammatical, but another is ungrammatical.
    
    Please solve the task by interleaving Thought, Action, and Answer steps. 
    Thought can reason about the current situation, and Action can be the following two types:
    
    (1) Follow-up Question[question], which returns a sub-question with a single answer that helps solve the original question.
    
    (2) Finish[], which means no more sub-questions. The final answer should be generated in the following line.
    
    [Demonstration Question]: Which sentence of the following two sentences is grammatical?
    
    FirstSentence[No author that no senators liked has had any success.]
    
    SecondSentence[The author that no senators liked has had any success.]
    
    [Thought 1]: Both of the two sentences use the word "any". in their most common uses, it can only be used in an appropriate syntactic-semantic-environment.
    
    [Action 1]: Follow-up Question[In what syntactic-semantic-environment can the word "any" be used?]
    
    [Answer 1]: To a first approximation, it can be only in the scope of negation.
    
    [Thought 2]: If a sentence does not contain a negation structure to match the word "any", it will be ungrammatical.
    
    [Action 2]: Follow-up Question[Which sentence does not contain a negation structure?]
    
    [Answer 2]: SecondSentence. Although it contains a negation structure of "no senators" in the subordinate clause, it does not contain a negation structure in the main clause to match the word "any" in the main clause.
    
    [Thought 3]: The SecondSentence of minimal pairs lacks a negation structure, so it is ungrammatical.
    
    [Action 3]: Finish[]

    [Answer 3]: FirstSentence is grammatical and SecondSentnce is ungrammatical.

    [More Demonstration Questions] [...]
    
    [Input Question]:
\end{tcolorbox}
\caption{Full version example of the capability-specific instruction.}
\label{fig:instruction-example-2}
\end{figure*}

\end{document}